\pdfoutput=1

\documentclass[11pt]{article}

\usepackage[]{acl}

\usepackage{times}
\usepackage{latexsym}
\usepackage{url}
\usepackage{paralist}
\usepackage{multirow}
\usepackage{verbatim}
\usepackage{amsfonts}
\usepackage{amssymb}
\usepackage{amsmath}
\usepackage{algorithm}
\usepackage{algpseudocode}
\usepackage{graphics}
\usepackage{epsfig}
\usepackage{multirow}
\usepackage{makecell}
\usepackage{booktabs}
\usepackage{float}

\usepackage{tabularx}
\newcolumntype{Y}{>{\centering\arraybackslash}X}

\newcommand{\StatexIndent}[1][3]{%
  \setlength\@tempdima{\algorithmicindent}%
  \Statex\hskip\dimexpr#1\@tempdima\relax}

\usepackage[T1]{fontenc}

\usepackage[utf8]{inputenc}

\usepackage{microtype}

\title{TREND: Trigger-Enhanced Relation Extraction Network for Dialogues}

\author{Po-Wei Lin\quad Shang-Yu Su\quad Yun-Nung Chen \\
National Taiwan University, Taipei, Taiwan\\
\texttt{\{r09922a24, f05921117\}@csie.ntu.edu.tw\quad y.v.chen@ieee.org}
}

\begin{document}
\maketitle
\begin{abstract}
The goal of dialogue relation extraction (DRE) is to identify the relation between two entities in a given dialogue.
During conversations, speakers may expose their relations to certain entities by explicit or implicit clues, such evidences called ``triggers''.
However, trigger annotations may not be always available for the target data, so it is challenging to leverage such information for enhancing the performance.
Therefore, this paper proposes to learn how to identify triggers from the data with trigger annotations and then transfers the trigger-finding capability to other datasets for better performance.
The experiments show that the proposed approach is capable of improving relation extraction performance of unseen relations and also demonstrate the transferability of our proposed trigger-finding model across different domains and datasets.\footnote{The source code is available at: \url{http://github.com/MiuLab/TREND}.}


\end{abstract}

\section{Introduction}

The goal of relation extraction (RE) is to identify the semantic relation type between two mentioned entities from a given text piece, which is one of basic and important natural language understanding (NLU) problems \cite{zhang2017position, zhou2021improved, cohen2020relation}.
In this task setting, we are usually given a written sentence and a query pair containing two entities and asked to return the most possible relation type from a predefined set of relations.
Dialogue relation extraction (DRE), on the other hand, aims to excavate underlying cross-sentence relation in natural human communications \cite{yu2020dialogue, jia2021ddrel}.
The problem itself is well-motivated, because relations between entities in dialogues could potentially provide dialogue systems with additional features for better dialogue managing~\cite{peng2018deep, su2018discriminative} or response generation~\cite{su2018natural}.  

\begin{figure}[t!]
\centering 
\includegraphics[width=\linewidth]{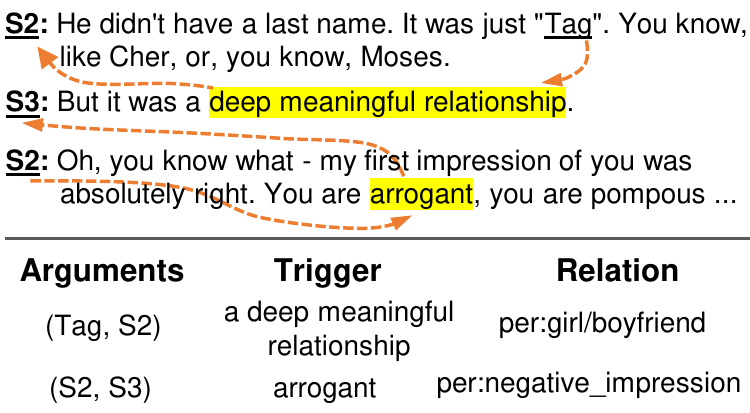}
\vspace{-5mm}
\caption{An example of dialogue relation extraction; the dashed arrows connect subjects, triggers, and objects. Triggers are clues of relations annotated in DialogRE.} 
\label{fig:example} 
\vspace{-3mm}
\end{figure}

\begin{figure*}[t!]
\centering 
\includegraphics[width=0.88\linewidth]{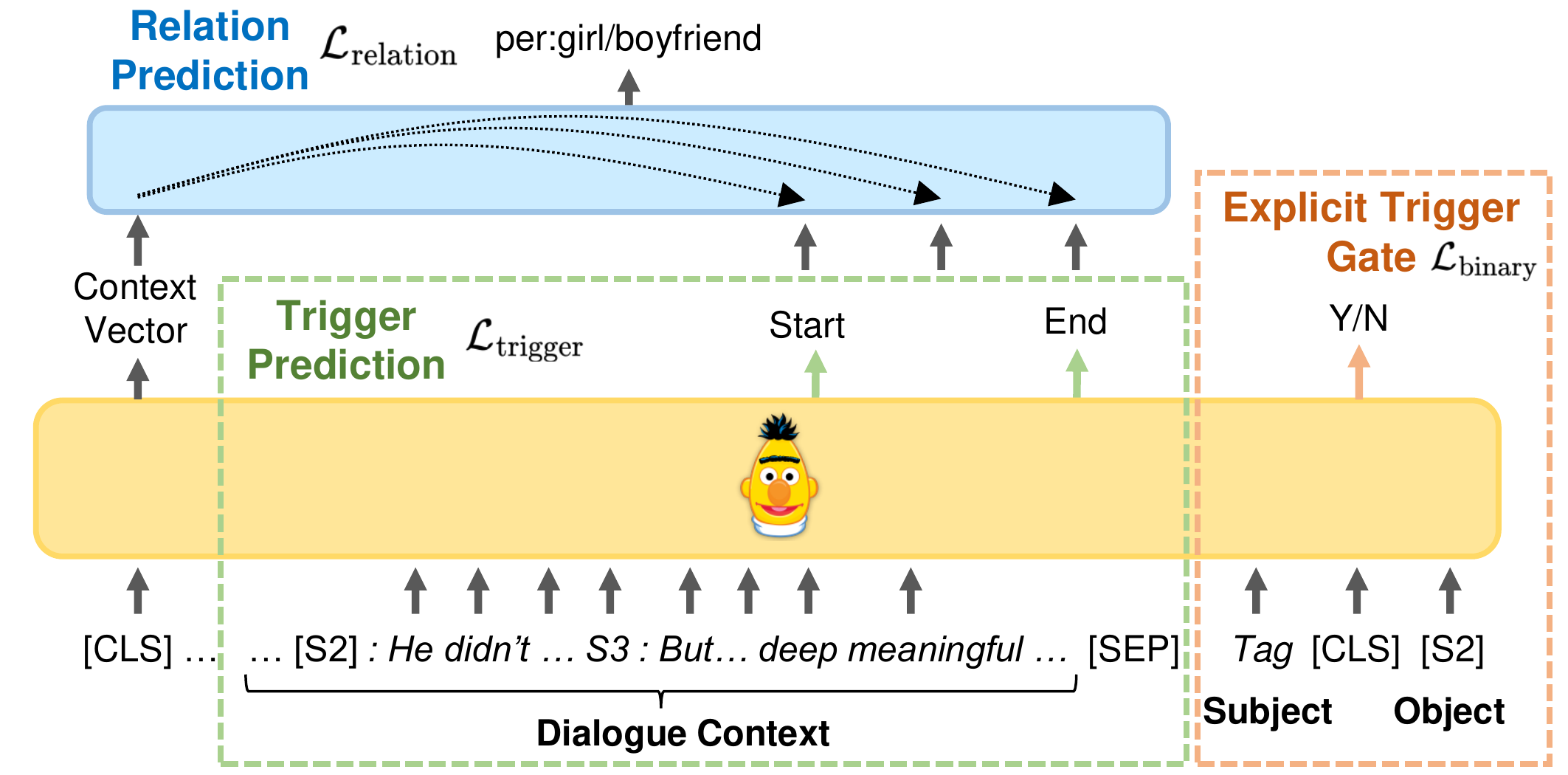}
 \vspace{-1mm}
\caption{The proposed method contains two components: (1) a multi-tasking BERT with two fine-tuning tasks (explicit trigger classification and trigger prediction), and (2) a relation predictor with attentional feature fusion.} 
\label{fig:framework} 
\vspace{-2mm}
\end{figure*}

There are two popular datasets, DialogRE~\cite{yu2020dialogue} and DDRel~\cite{jia2021ddrel}, focusing on relation extraction in dialogues illustrated in Figure \ref{fig:example}.
In DRE, given a conversation and a query pair, we aim to identify the interpersonal relationship between the given entities, where entities can be human or other types like locations.
As shown in Figure~\ref{fig:example}, the evidences of relations within the conversation flow, called \textbf{Triggers}, provide informative cues for this task.
A trigger can be a short phrase or even a single word with any possible part-of-speech.
In the example, the clue for knowing the speaker 2 has a negative impression on the speaker 3 comes from the sentence ``\emph{You are arrogant.}''
Such hint is intuitively useful for deciding the relations.
However, \citet{albalak2022d} is the only prior work that tried to explicitly leverage such signal for improving DRE, because such explanation annotations may not be always available~\cite{kung2020zero}.

Prior work can be divided into two main lines, one of which is graph-based methods.
DHGAT \cite{chen2020dialogue} presents an attention-based heterogeneous graph network to model multiple types of features;
GDPNet \cite{xue2021gdpnet} constructs latent multi-view graphs to model possible relationships among tokens in a long sequence, and then refines the graphs by iterative graph convolution and pooling techniques.
Another branch is BERT-based \cite{kenton2019bert} methods \cite{yu2020dialogue, xue2022embarrassingly}. 
SimepleRE \cite{xue2022embarrassingly} is a simple BERT model with an additional refinement gate for iteratively finding high-confidence prediction.
LSR \cite{nan2020reasoning} is a latent structure refinement method for better reasoning in the document-level relation extraction task.
Although it is known that using trigger information can significantly help the performance of relation extraction, only DialogRE has the annotated triggers.
It is not guaranteed that utilizing the annotated triggers can generalize to other relations from other datasets, considering the discrepancy of their relation types.

Given the target data without trigger annotations, this paper proposes \textbf{TREND}, a simple multi-tasking model with an attentional relation predictor, 
where it learns the general capability of finding triggers and transfers it to the unseen relations for performance improvement.
The experiments show that our proposed method can effectively identify the explicit triggers and generalize to unseen relations towards great flexibility and practicality.


\section{Proposed Method}
The core idea of this model is to identify trigger spans and accordingly leverage such signal to improve relation extraction. 
We hereby propose \textbf{T}rigger-enhanced \textbf{R}elation-\textbf{E}xtraction \textbf{N}etwork for \textbf{D}ialogues, \textbf{TREND}, illustrated in Figure~\ref{fig:framework}.

\subsection{Problem Formulation}
Given a piece of dialogue context $\mathcal{D}$ composed of text tokens $\mathcal{D} = \{ x_i \}$ and a query pair $q$ containing a subject entity and an object entity $q = (s, o)$, the task aims at learning a function $f$ that finds the most possible relations between the given entities from a predefined relation set $\mathcal{R}$,
$f (\mathcal{D}, q) \to \mathcal{R}.$
Note that a single query pair may contain multiple relations, and we duplicate the data samples when they have multiple relation labels by following the prior work.

\subsection{TREND}
The proposed model has two modules, (1) a multi-tasking BERT \cite{kenton2019bert} for encoding context and identifying triggers, and (2) a relation predictor with a feature fusion of the dialogue and the automatically identified trigger.

As illustrated in Figure \ref{fig:framework}, an input $(\mathcal{D}, q)$ will be first augmented into a BERT-style sequence.
Specifically, the input format is ``$\texttt{[CLS]} \ \mathcal{D} \ \texttt{[SEP]} \ s \ \texttt{[CLS]} \ o$''.
We replace the target entity pair with their speaker tokens in $\mathcal{D}$ following \citet{yu2020dialogue} illustrated in the figure.
The first \texttt{[CLS]} encodes the dialogue contexts, and the second one is to predict whether the triggers are explicit via binary classification detailed below.

\vspace{-1mm}
\paragraph{Explicit Trigger Gate}
Because triggers sometimes are implicit, it is difficult to identify the associated trigger spans of dialogue relations.
We hereby propose to learn a binary classifier as a gate to identify if the explicit triggers exist, and empty trigger spans are inputted to relation prediction when no explicit triggers. 
The binary cross entropy loss $\mathcal{L}_\text{binary}$ is used here. 

\vspace{-1mm}
\paragraph{Trigger Prediction}
The explicit triggers are identified by a extractive method with start-end pointer prediction~\cite{kenton2019bert}, which is prevalent in extractive question answering~\cite{lee2016learning, Rajpurkar_2016}.
This is a single-label classification problem of predicting the most possible positions; hence a cross entropy loss $\mathcal{L}_\text{trigger}$ is conducted.


\vspace{-1mm}
\paragraph{Relation Prediction}
A learned context vector and a predicted trigger span are then fed into the relation predictor as depicted in the top part of Figure \ref{fig:framework}.
The features are fused by a generic attention mechanism, where the query is the context vector $\mathbf{c}$, and the keys and the values are trigger words $\mathbf{x_i}$ encoded by BERT:
$\sum \mathrm{softmax}(\mathbf{c} \cdot \mathbf{x_i}) \cdot \mathbf{x_i}$.
The merged feature is then fed into a 1-layer feed-forward network for final relation prediction using a cross entropy loss $\mathcal{L}_\text{relation}$.

\vspace{-1mm}
\paragraph{Supervised Joint Learning}
Considering that only DialogRE contains the annotated trigger cues, we perform supervised joint learning for three above tasks.
Three above losses are linearly combined as the learning objective for training the whole model in an end-to-end manner.
The weights for adjusting the impact of each loss are tuned in the development set.
We also apply schedule sampling \cite{bengio2015scheduled} on explicit trigger classification and trigger prediction when feeding into the relation predictor in order to mitigate the gap between the true triggers and the predicted ones.

\vspace{-1mm}
\paragraph{Transfer Learning}
Because annotated triggers may not be available, this paper focuses on transferring the trigger-finding capability to another target dataset, DDRel, which does not contain trigger annotations and its relation types differ a lot from DialogRE.
We replace the final feed-forward layer with a new one, since relation numbers may different in two datasets.
Then we fine-tune the whole model using a single loss about relation prediction, $\mathcal{L}_\text{relation}$, where we assume the trigger-finding capability can be better transferred cross different datasets/relations.



\begin{table}[t!]
\centering
\begin{tabular}{ | l | c |}
    \hline
    \bf  Model & \bf F1\\
\hline \hline
 BERT  & 60.6 \\
 GDPNet & 64.3\\
 SimpleRE (single entity pair)  & 60.4 \\
 D-REX$_\text{BERT}$ & 59.2 \\
 TUCORE-GCN$_\text{BERT}$ & 65.5  \\
  \bf TREND$_\text{BERT-Base}$  & \bf 66.8 \\
 \bf TREND$_\text{BERT-Large}$  & \bf 67.8 \\
\hline
SimpleRE (multiple entity pairs) & 66.7 \\
SocAoG (multiple entity pairs) & \it 69.1 \\
\hline
TREND$_\text{BERT-Base}$ (ground-truth triggers) & 75.3\\
\hline
 
\end{tabular}
\vspace{-1mm}
\caption{The model performance on DialogRE.}
\label{tab:dialogre}
\vspace{-3mm}
\end{table}

\section{Experiments}
\label{sec:exp}

We focus on evaluating the performance of DRE on the dataset without trigger labels in order to investigate if the trigger-finding capability can be transferred across datasets/relations.

\subsection{Setting}
The DRE datasets used in our experiments are DialogRE (v2) with trigger annotations \cite{yu2020dialogue} and DDRel \cite{jia2021ddrel} without trigger annotations.
Text normalization like lemmatization and expanding contractions is applied to data preprocessing.
In all experiments, we use mini-batch \texttt{adam} with a learning rate $3\mathrm{e}{-5}$ as the optimizer on Nvidia Tesla V100.
The ratio of teacher forcing and other hyper-parameters are selected by grid search in (0,1] with a step 0.1.
The training takes 30 epochs without early stop.
The detailed implementation can be found in Appendix \ref{sec:details}.

The following BERT-based methods are performed for fair comparison:
1) BERT, 2) GDPNet~\cite{xue2021gdpnet}, 3) SimpleRE~\cite{xue2022embarrassingly}, 4) D-REX$_\text{BERT}$~\cite{albalak2022d}, and 5)  TUCORE-GCN$_\text{BERT}$~\cite{lee2021graph}.
Other approaches that take multiple entity pairs for global consideration cannot directly be compared with TREND but reported as reference.

\begin{table*}[t!]
\centering
\begin{tabular}{ | l | c c c c c c| }
    \hline
    \multirow{2}{*}{\bf Model}  & \multicolumn{2}{c}{\bf 4-class} & \multicolumn{2}{c}{\bf  6-class} & \multicolumn{2}{c|}{\bf 13-class}  \\
         & \bf Acc & \bf Macro-F & \bf Acc & \bf Macro-F  & \bf Acc & \bf Macro-F  \\
\hline \hline
BERT  & 47.1 / 58.1 & 44.5 / 52.0 & 41.9 / 42.3 & 39.4 / 38.0 & 39.4 / 39.7 & 20.4 / 24.1 \\
TUCORE-GCN$_\text{BERT}$ & 43.8 / 60.3 & 41.9 / 56.6 & 36.9 / \textbf{52.6} & 38.7 / 54.2 & 29.5 / 44.9 & 20.5 / \textbf{36.9}\\
\hline
TREND$_\text{BERT-Base}$  & 51.5 / \textbf{65.4} & \bf 46.5 / 61.2 & 40.3 / \textbf{52.6} & \bf 43.0 / 55.0 & \bf 40.5 / 46.2 & 21.2 / 34.7 \\
\ \ \ w/o binary gate & 52.5 / 53.8 & 45.3 / 49.7 & 37.0 / 43.6 & 41.8 / 45.9 & 36.6 / 43.6 &  \textbf{26.4} / 36.3 \\
TREND$_\text{BERT-Large}$ & \textbf{51.6} / 60.3 & \textbf{46.5} / 54.0 & \textbf{42.5} / 46.2 & \textbf{43.0} / 48.2 & 34.4 / 43.6 & 19.9 / 36.3 \\
\ \ \ w/o binary gate & 41.5 / 47.4 & 40.3 / 44.9 & 39.0 / 42.3 & 43.1 / 42.9 & 38.5 / 34.6 & 17.3 / 21.1 \\
\hline
\end{tabular}
\vspace{-1mm}
\caption{The DDRel performance in session-level/pair-level settings and different granularity settings (4,6,13-class).
}
\vspace{-3mm}
\label{tab:ddrel}
\end{table*}

\subsection{Results of Supervised Joint Learning}
The performance of our TREND model jointly trained on the trigger-available DialogRE dataset is presented in Table~\ref{tab:dialogre}, where it is obvious that our TREND achieves the best performance in the fair setting.
Unlike SimpleRE and GDPNet that need to iteratively refine the latent features or latent graphs, relation prediction in the proposed TREND is straight-forward, making training and inference efficient and robust.
Furthermore, D-REX~\cite{albalak2022d} also leverages triggers for relation prediction but performs significantly worse than our simple TREND models in the same setting.
Our trained binary gate has about 85\% accuracy while the trigger prediction has no more than 50\% of exact match.
Although our model cannot perfectly extract the triggers, the predicted spans can still facilitate relation prediction in our proposed TREND.
It demonstrates that our TREND model is capable of identifying potential triggers and utilizing such cues for predicting relations.
Note that TREND$_\text{BERT-Large}$ is for reference, indicating that a larger model has the potential of further improving the performance.
The upper-bound of our proposed TREND$_\text{BERT-Base}$ is 75.3 shown in the last row of Table~\ref{tab:dialogre}, where the ground truth triggers are inputted in the relation predictor. 
This higher score suggests that our TREND model still has a room for improvement and the proposed model design is well validated.

\subsection{Results of Transfer Learning}
Due to the lack of trigger annotations in DDRel, our TREND model focuses on transferring the trigger-finding capability learned from DialogRE to DDRel.
We compare our proposed TREND with two models, which are not designed for transferring across different relation extraction datasets, so they are directly trained on the DDRel data.
Table~\ref{tab:ddrel} presents the performance achieved on DDRel evaluated in session-level and pair-level settings, where session-level relation extraction is given a \emph{partical} dialogue the entity pair is involved in and pair-level is based on a \emph{full} dialogue~\cite{jia2021ddrel}.\footnote{A session only contains multiple turns in a dialogue, so session-level results are worse than pair-level ones.}
All scores are much lower than ones in DialogRE due to the higher difficulty of this dataset.
The obtained improvement compared with the BERT baseline is larger when the longer dialogue contexts as the input; that is, pair-level improvement is more than  session-level one.
The probable reason is that extracting key evidences for predicting relations is more important to overcome information overload.

Furthermore, we report the performance of the current state-of-the-art (SOTA) relation extraction model, TUCORE-GCN, on the DDRel dataset.\footnote{The numbers are obtained based on the released code in \citet{lee2021graph}.}
It can be found that our proposed method can effectively transfer the capability of capturing triggers from DialogRE to DDRel, and outperform TUCORE-GCN in most cases, achieving a new SOTA performance in DDRel.

Suprisingly, TREND$_\text{BERT-Large}$ does not outperform TREND$_\text{BERT-Base}$, implying that TREND$_\text{BERT-Base}$ already has enough good capability of capturing triggers and can generalize to another dataset (DDRel) and a new relation set.

\subsection{Ablation Study}
Because our trigger finding module contains a binary classifier deciding the existence of explicit triggers and a trigger predictor extracting trigger spans, we examine the effectiveness of the binary gate.
By removing the binary gate, the performance is consistently degraded shown in Table~\ref{tab:ddrel}, further demonstrating the effectiveness of the designed trigger-finding module in our TREND model.

\subsection{Generalization of Unseen Relations}
To further investigate if our trigger-finding capability can generalize to different relations, we categorize all relations into seen and unseen relations based on the relation similarity between the two datasets shown in Table~\ref{tab:ontology}, and show the session-level performance in Table~\ref{tab:unseen}.
It can be seen that our proposed TREND is capable of transferring trigger-finding capability from DialogRE to DDRel, even DDRel does not contain trigger annotations.
More importantly, our learned trigger-finding capability is demonstrated general to diverse relations, because TREND achieves better results for not only seen but also unseen relations whose triggers never appear in the DialogRE data.
We qualitatively analyze the predicted triggers of unseen relations, where TREND extracts a dirty word (``fxxk'') and a word ``client'' as triggers for unseen relations ``opponent'' and ``professional contact'' in DDRel respectively.
The full samples can be found in Table~\ref{tab:example_ddrel}.
It shows the effectiveness and generalizability of our proposed TREND model towards practical usage.

\begin{table}[t!]
    \centering
    \small
    \begin{tabular}{ll}
    \toprule
    \bf DDRel Relation & \bf DialogRE Relation\\
    \midrule
     Workplace Superior-Subordinate & per:boss \\
     Workplace Superior-Subordinate & per:subordinate\\
     Friends & per:friends\\
     Lovers & per:girl/boyfriend\\
     Neighbors & per:neighbor\\
     Roommates & per:roommate\\
     Child-Parent & per:children\\
     Child-Other Family Elder & per:other family\\
     Siblings & per:siblings\\
     Spouse & per:spouse\\
     Colleague/Partners & per:works\\
    \midrule
     Courtship & -\\
     Opponents & -\\
     Professional Contact & -\\
    \bottomrule
    \end{tabular}
    \vspace{-1mm}
    \caption{Relation ontology mapping between DDRel and DialogRE datasets.}
    \label{tab:ontology}
    \vspace{-2mm}
\end{table}

\begin{table}[t]
    \centering
    \begin{tabular}{|l|cc|}
    \hline
    \bf DDRel Relation & \bf Seen & \bf Unseen \\
    \hline\hline
    BERT & 23.77 & ~~9.94 \\
    TUCORE-GCN & 23.39 & 10.81 \\
    TREND & 28.30 & 13.13 \\
    \hline
    \end{tabular}
    \vspace{-1mm}
    \caption{F1 results of DDRel seen and unseen relations.}
    \label{tab:unseen}
    \vspace{-2mm}
\end{table}

\subsection{Qualitative Study}
\label{sec:qual}
The predicted triggers and relation for DialogRE and DDRel datasets are presented in Table~\ref{tab:example_dialogre} and Table \ref{tab:example_ddrel} respectively.
Note that the triggers are not annotated in DDRel.
It can be found that TREND can extract explicit cues as triggers not only for the seen relations, which are similar to relations in DialogRE, but also unseen ones.

\section{Conclusion}
This paper proposes TREND, a multi-tasking model with the generalizable trigger-finding capability, to improve dialogue relation extraction.
TREND is a simple, flexible, end-to-end model based on BERT with three components:
(1) an explicit trigger gate for trigger existence,
(2) an extractive trigger predictor,
and (3) a relation predictor with an attentional feature fusion.
The experiments demonstrate that TREND can successfully transfer the learned trigger-finding capability across different datasets and diverse relations for better dialogue relation extraction performance, showing the great potential of improving explainability without rationale annotations.

\begin{table}[t!]
 \small
\begin{tabular}{  | p{2cm}|p{2.6cm} | p{1.8cm} | }
\hline
\multicolumn{3}{|p{7.3cm}|}{
S1: What's up?
} \\
\multicolumn{3}{|p{7.3cm}|}{
S2: Monica and I are \textbf{engaged}.
} \\
\multicolumn{3}{|p{7.3cm}|}{
S1: Oh my God. Congratulations.
} \\
\multicolumn{3}{|p{7.3cm}|}{
S2: Thanks.
} \\
\hline
\bf Argument & \bf Relation  & \bf Trigger \\
 (S2, Monica) &  girl/boyfriend &  engaged \\
 
    \hline
  \end{tabular}
  \vspace{-1mm}
\caption{A predicted result of TREND on DialogRE. }
\label{tab:example_dialogre}
 \vspace{-2mm}
\end{table}


\begin{table}[t!]
\small
\begin{tabular}{  | p{2cm}|p{2.6cm} | p{1.8cm} | }
\hline
\multicolumn{3}{|p{7.3cm}|}{
S1: That's all. 
} \\
\multicolumn{3}{|p{7.3cm}|}{
S2: That's all?!
} \\
\multicolumn{3}{|p{7.3cm}|}{
S1: You don't see it, do you, \textbf{father}?"
} \\
\multicolumn{3}{|p{7.3cm}|}{
S2: No. Fellow wants to sell a house ...
} \\
    \hline
    \bf \small Argument & \bf Relation (Seen) & \bf Trigger \\
     (S1, S2) & \small Child-Parent & \small father \\
    \hline\hline
\multicolumn{3}{|p{7.3cm}|}{
S1: \textbf{Fuck} me! 
} \\
\multicolumn{3}{|p{7.3cm}|}{
S2: Want a drink? Okay...
I'm not good at this sort of thing, but we don't have a lot of time, so I'll just go ahead and get started.
} \\
\hline
\bf  Argument & \bf  Relation (Unseen) & \bf Trigger \\
(S1, S2) &  Opponent & fuck \\
\hline\hline
\multicolumn{3}{|p{7.3cm}|}{
S1: I'm Joe Galvin, I'm representing Deborah Ann Kaye, case against St.
} \\
\multicolumn{3}{|p{7.3cm}|}{
S2: I told the guy I didn't want to talk to...
} \\
\multicolumn{3}{|p{7.3cm}|}{
S1: I'll just take a minute. Deborah Ann Kaye. You know what I'm talking about.
} \\
\multicolumn{3}{|p{7.3cm}|}{
S2: No.
} \\
\multicolumn{3}{|p{7.3cm}|}{
S1: He's the Assistant Chief of Anesthesiology, Massachusetts Commonwealth. He says your doctors, Towler and Marx, put my girl in the hospital for life. And we can prove that. What we don't know is why. I want someone who was in the O.R.
} \\
\multicolumn{3}{|p{7.3cm}|}{
S2: I've got nothing to say to you.
} \\
\multicolumn{3}{|p{7.3cm}|}{
S1: You know what happened.
} \\
\multicolumn{3}{|p{7.3cm}|}{
S2: Nothing happened.
} \\
\multicolumn{3}{|p{7.3cm}|}{
S1: Then why aren't you testifying for their side? I can subpoena you, you know. I can get you up there on the stand.
} \\
\multicolumn{3}{|p{7.3cm}|}{
S2: And ask me what?
} \\
\multicolumn{3}{|p{7.3cm}|}{
S1: Who put my \textbf{client} in the hospital for life.
} \\
\multicolumn{3}{|p{7.3cm}|}{
S2: I didn't do it, Mister.
} \\
\multicolumn{3}{|p{7.3cm}|}{
S1: Who are you protecting, then?
} \\
\multicolumn{3}{|p{7.3cm}|}{
S2: Who says that I'm protecting anyone?
} \\
\multicolumn{3}{|p{7.3cm}|}{
S1: I do. Who is it? The Doctors. What do you owe them?
} \\
\multicolumn{3}{|p{7.3cm}|}{
S2: I don't owe them a goddamn thing.
} \\
\multicolumn{3}{|p{7.3cm}|}{
S1: Then why don't you testify?
} \\
\multicolumn{3}{|p{7.3cm}|}{
S2: You know, you're pushy, fella...
} \\
\multicolumn{3}{|p{7.3cm}|}{
S1: You think I'm pushy now, wait 'til I get you on the stand...
} \\
\multicolumn{3}{|p{7.3cm}|}{
S2: Well, maybe you better do that, then.
} \\
\hline
\bf  Argument & \bf  Relation (Unseen) & \bf Trigger \\
(S1, S2) & Professional Contact & client \\
\hline
  \end{tabular}
  \vspace{-1mm}
\caption{Predicted results of TREND on DDRel.}
 \vspace{-2mm}
\label{tab:example_ddrel}
\end{table}

\section*{Acknowledgements}
We thank reviewers for their insightful comments and Ze-Song Xu for running baselines.
This work was financially supported from Google and the Young Scholar Fellowship Program by Ministry of Science and Technology (MOST) in Taiwan, under Grants 111-2628-E-002-016 and 111-2634-F-002-014.

\bibliography{anthology,custom}

\begin{thebibliography}{19}
\expandafter\ifx\csname natexlab\endcsname\relax\def\natexlab#1{#1}\fi

\bibitem[{Albalak et~al.(2022)Albalak, Embar, Tuan, Getoor, and
  Wang}]{albalak2022d}
Alon Albalak, Varun Embar, Yi-Lin Tuan, Lise Getoor, and William~Yang Wang.
  2022.
\newblock {D-REX}: Dialogue relation extraction with explanations.
\newblock In \emph{Proceedings of the 4th Workshop on NLP for Conversational
  AI}, pages 34--46.

\bibitem[{Bengio et~al.(2015)Bengio, Vinyals, Jaitly, and
  Shazeer}]{bengio2015scheduled}
Samy Bengio, Oriol Vinyals, Navdeep Jaitly, and Noam Shazeer. 2015.
\newblock Scheduled sampling for sequence prediction with recurrent neural
  networks.
\newblock In \emph{Advances in Neural Information Processing Systems}, pages
  1171--1179.

\bibitem[{Chen et~al.(2020)Chen, Hong, Han, Majumder, and
  Poria}]{chen2020dialogue}
Hui Chen, Pengfei Hong, Wei Han, Navonil Majumder, and Soujanya Poria. 2020.
\newblock Dialogue relation extraction with document-level heterogeneous graph
  attention networks.
\newblock \emph{arXiv preprint arXiv:2009.05092}.

\bibitem[{Cohen et~al.(2020)Cohen, Rosenman, and Goldberg}]{cohen2020relation}
Amir~DN Cohen, Shachar Rosenman, and Yoav Goldberg. 2020.
\newblock Relation classification as two-way span-prediction.
\newblock \emph{arXiv preprint arXiv:2010.04829}.

\bibitem[{Jia et~al.(2021)Jia, Huang, and Zhu}]{jia2021ddrel}
Qi~Jia, Hongru Huang, and Kenny~Q Zhu. 2021.
\newblock Ddrel: A new dataset for interpersonal relation classification in
  dyadic dialogues.
\newblock In \emph{Proceedings of the AAAI Conference on Artificial
  Intelligence}, volume~35, pages 13125--13133.

\bibitem[{Kenton and Toutanova(2019)}]{kenton2019bert}
Jacob Devlin Ming-Wei~Chang Kenton and Lee~Kristina Toutanova. 2019.
\newblock {BERT}: Pre-training of deep bidirectional transformers for language
  understanding.
\newblock In \emph{Proceedings of NAACL-HLT}, pages 4171--4186.

\bibitem[{Kung et~al.(2020)Kung, Yang, Chen, Yin, and Chen}]{kung2020zero}
Po-Nien Kung, Tse-Hsuan Yang, Yi-Cheng Chen, Sheng-Siang Yin, and Yun-Nung
  Chen. 2020.
\newblock Zero-shot rationalization by multi-task transfer learning from
  question answering.
\newblock In \emph{Findings of the Association for Computational Linguistics:
  EMNLP 2020}, pages 2187--2197.

\bibitem[{Lee and Choi(2021)}]{lee2021graph}
Bongseok Lee and Yong~Suk Choi. 2021.
\newblock Graph based network with contextualized representations of turns in
  dialogue.
\newblock In \emph{Proceedings of the 2021 Conference on Empirical Methods in
  Natural Language Processing}, pages 443--455.

\bibitem[{Lee et~al.(2016)Lee, Salant, Kwiatkowski, Parikh, Das, and
  Berant}]{lee2016learning}
Kenton Lee, Shimi Salant, Tom Kwiatkowski, Ankur Parikh, Dipanjan Das, and
  Jonathan Berant. 2016.
\newblock Learning recurrent span representations for extractive question
  answering.
\newblock \emph{arXiv preprint arXiv:1611.01436}.

\bibitem[{Nan et~al.(2020)Nan, Guo, Sekuli{\'c}, and Lu}]{nan2020reasoning}
Guoshun Nan, Zhijiang Guo, Ivan Sekuli{\'c}, and Wei Lu. 2020.
\newblock Reasoning with latent structure refinement for document-level
  relation extraction.
\newblock In \emph{Proceedings of the 58th Annual Meeting of the Association
  for Computational Linguistics}, pages 1546--1557.

\bibitem[{Peng et~al.(2018)Peng, Li, Gao, Liu, Wong, and Su}]{peng2018deep}
Baolin Peng, Xiujun Li, Jianfeng Gao, Jingjing Liu, Kam-Fai Wong, and Shang-Yu
  Su. 2018.
\newblock Deep dyna-q: Integrating planning for task-completion dialogue policy
  learning.
\newblock \emph{arXiv preprint arXiv:1801.06176}.

\bibitem[{Rajpurkar et~al.(2016)Rajpurkar, Zhang, Lopyrev, and
  Liang}]{Rajpurkar_2016}
Pranav Rajpurkar, Jian Zhang, Konstantin Lopyrev, and Percy Liang. 2016.
\newblock \href {https://doi.org/10.18653/v1/d16-1264} {Squad: 100,000+
  questions for machine comprehension of text}.
\newblock \emph{Proceedings of the 2016 Conference on Empirical Methods in
  Natural Language Processing}.

\bibitem[{Su et~al.(2018{\natexlab{a}})Su, Li, Gao, Liu, and
  Chen}]{su2018discriminative}
Shang-Yu Su, Xiujun Li, Jianfeng Gao, Jingjing Liu, and Yun-Nung Chen.
  2018{\natexlab{a}}.
\newblock Discriminative deep dyna-q: Robust planning for dialogue policy
  learning.
\newblock In \emph{Proceedings of the 2018 Conference on Empirical Methods in
  Natural Language Processing}, pages 3813--3823.

\bibitem[{Su et~al.(2018{\natexlab{b}})Su, Lo, Yeh, and Chen}]{su2018natural}
Shang-Yu Su, Kai-Ling Lo, Yi-Ting Yeh, and Yun-Nung Chen. 2018{\natexlab{b}}.
\newblock Natural language generation by hierarchical decoding with linguistic
  patterns.
\newblock In \emph{Proceedings of The 16th Annual Conference of the North
  American Chapter of the Association for Computational Linguistics: Human
  Language Technologies}.

\bibitem[{Xue et~al.(2021)Xue, Sun, Zhang, and Chng}]{xue2021gdpnet}
Fuzhao Xue, Aixin Sun, Hao Zhang, and Eng~Siong Chng. 2021.
\newblock {GDPN}et: Refining latent multi-view graph for relation extraction.
\newblock In \emph{Proceedings of the AAAI Conference on Artificial
  Intelligence}, volume~35, pages 14194--14202.

\bibitem[{Xue et~al.(2022)Xue, Sun, Zhang, Ni, and
  Chng}]{xue2022embarrassingly}
Fuzhao Xue, Aixin Sun, Hao Zhang, Jinjie Ni, and Eng-Siong Chng. 2022.
\newblock An embarrassingly simple model for dialogue relation extraction.
\newblock In \emph{ICASSP 2022-2022 IEEE International Conference on Acoustics,
  Speech and Signal Processing (ICASSP)}, pages 6707--6711. IEEE.

\bibitem[{Yu et~al.(2020)Yu, Sun, Cardie, and Yu}]{yu2020dialogue}
Dian Yu, Kai Sun, Claire Cardie, and Dong Yu. 2020.
\newblock Dialogue-based relation extraction.
\newblock In \emph{Proceedings of the 58th Annual Meeting of the Association
  for Computational Linguistics}, pages 4927--4940.

\bibitem[{Zhang et~al.(2017)Zhang, Zhong, Chen, Angeli, and
  Manning}]{zhang2017position}
Yuhao Zhang, Victor Zhong, Danqi Chen, Gabor Angeli, and Christopher~D Manning.
  2017.
\newblock Position-aware attention and supervised data improve slot filling.
\newblock In \emph{Proceedings of the 2017 Conference on Empirical Methods in
  Natural Language Processing}, pages 35--45.

\bibitem[{Zhou and Chen(2021)}]{zhou2021improved}
Wenxuan Zhou and Muhao Chen. 2021.
\newblock An improved baseline for sentence-level relation extraction.
\newblock \emph{arXiv preprint arXiv:2102.01373}.

\end{thebibliography}
\bibliographystyle{acl_natbib}

\appendix

\section{Reproducibility}
\label{sec:details}

\subsection{Hyperparameters}
All the hyper-parameters were selected by grid search in (0,1] with step 0.1.
The loss functions are linearly combined and each of them has an adjustable weight.

\paragraph{TREND$_\text{BERT-Base}$}
\begin{compactitem}
\item Loss: $0.3 \cdot \mathcal{L}_\text{trigger}$ + $1.0 \cdot \mathcal{L}_\text{relation}$ + $1.0 \cdot \mathcal{L}_\text{binary}$
\item schedule sampling: 0.7 for trigger prediction, 0.7 for binary classification
\end{compactitem}

\paragraph{TREND$_\text{BERT-Large}$}
\begin{compactitem}
\item Loss: $0.3 \cdot \mathcal{L}_\text{trigger}$ + $1.0 \cdot \mathcal{L}_\text{relation}$ + $1.0 \cdot \mathcal{L}_\text{binary}$
\item schedule sampling: 0.5 for trigger prediction, 0.7 for binary classification
\end{compactitem}

\subsection{Time Efficiency}
The training and inference cost in terms of time is reported in Table~\ref{tab:time}.
\begin{table}[t]
\centering
\resizebox{0.99\linewidth}{!}{
\begin{tabular}{ | l | c | c | }
    \hline
    \bf Data & \bf Training & \bf Inference \\
\hline \hline
DialogRE & 15 mins$\times$30 & 5 mins\\
DDRel (session-level) & 15 mins$\times$30 & 5 mins\\
DDRel (pair-level) & 1.5 mins$\times$30 & 10 secs\\
\hline
\end{tabular}}
\vspace{-2mm}
\caption{Time efficiency on three sets of experiments.}
\label{tab:time}
\vspace{-2mm}
\end{table}

\end{document}